\documentclass[letterpaper]{article} 
\usepackage{aaai2027}
\usepackage[hyphens]{url}  
\usepackage{graphicx} 
\usepackage{natbib}  
\usepackage{caption} 
\usepackage{algorithm}
\usepackage{algorithmic}
\usepackage{amssymb}
\usepackage{multirow}
\usepackage{amsmath}
\newtheorem{theorem}{Theorem}
\newtheorem{proposition}{Proposition}

\usepackage{newfloat}
\usepackage{listings}
\DeclareCaptionStyle{ruled}{labelfont=normalfont,labelsep=colon,strut=off} 
\floatstyle{ruled}
\newfloat{listing}{tb}{lst}{}
\floatname{listing}{Listing}

\usepackage{booktabs}

\title{Chebyshev Manifold Adaptation}
\author{
    Jiawen Li\textsuperscript{\rm 1}
}
\affiliations{
    \textsuperscript{\rm 1}School of Computer Science and Engineering,
    University of New South Wales, Sydney, Australia\\
    jiawen.li12@student.unsw.edu.au\\
    ORCID: 0009-0002-1449-2803
}

\begin{document}

\maketitle

\begin{abstract}
The paper presents a new parameter-efficient adaptation method called ChebyMA (Chebyshev Manifold Adaptation). ChebyMA adopts weight matrices through a multi-surface superposition of Chebyshev polynomial bases evaluated on learnable coordinates and combined via trainable coefficient matrices, replacing standard linear projections with highly expressive continuous function approximation. Theoretically, we establish an Approximation Expressivity Theorem, proving from the perspective of function approximation theory that single-manifold ChebyMA guarantees convergence in Frobenius norm error of reconstruction. Besides, drawing on Kolmogorov $n$-width intuition, we demonstrate the expressive advantages of multi-manifold superposition ($S > 1$) in decoupling high-dimensional complex features. Experimental results on Computer Vision CIFAR datasets(CIFAR-10, CIFAR-100)\cite{CIFAR} and Natural Language Processing (AG News, SST-2) datasets demonstrate that ChebyMA consistently achieves a superior parameter-accuracy Pareto front compared to standard full-parameter fine-tuning, LoRA\cite{LoRA}, TLoRA\cite{TLoRA}, and StelLA\cite{StelLA}. ChebyMA significantly outperforms other tested methods in tested datasets, validating its solid theoretical foundation for generality with purely vectorized computations.
\end{abstract}


\section{Introduction}
In recent years, Deep Learning models have been widely utilized in a wide range of applications, including image/text/audio recognition, generation, or real-time physical simulation. However, full model training usually costs a lot computation resources and leads to inconvenience for its usability. LoRA, also known as Low Rank Adaptation, tries to solve this issue by simply decomposing the parameter matrix need to be trained into two low-rank matrices, which largely reduces the time of this process. Albeit advantages of LoRA to the reduction of parameters, it and its subsequent variants (e.g., TLoRA, StelLA) inherently still have a bottleneck that is highly constrained by linear projections, which the objective is complex enough, these simple structures lack enough expressivity.

To overcome this problem,we rethink parameter-efficient fine-tuning in another aspect about function approximation. Instead of relying on linear matrix decomposition,the method we propose, ChebyMA, formulates the weights through a multi-surface superposition of Chebyshev polynomial bases. Our empirical designs rigidly grounded in mathematical theory about Approximation Expressivity Theorem rather than simply heuristic.

\begin{figure}[h]
\centering
\includegraphics[width=1\linewidth]{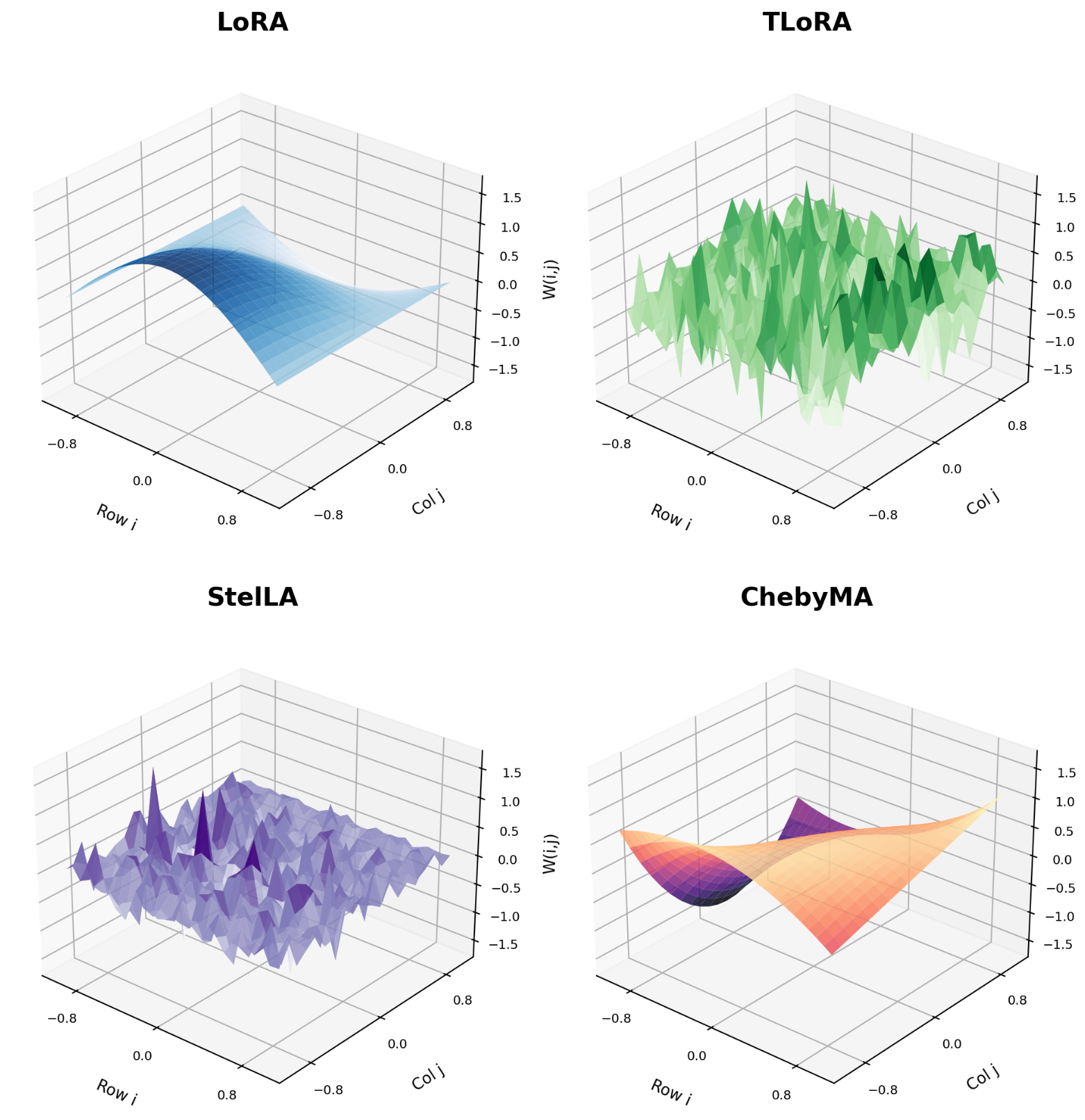}
\caption{Overview of different methods}
\label{fig:overview}
\end{figure}

The central insight is geometric: we treat each 
weight matrix as a scalar field over its 
row-column index space and approximate it through 
a superposition of Chebyshev polynomial 
hypersurfaces. Details are developed in the Method section.

This perspective also reveals a unifying view of existing methods. LoRA, TLoRA, and StelLA can all be understood as special cases within a three-factor decomposition framework, differing only in how their outer matrices are constructed: random and frozen (TLoRA), orthogonally constrained (StelLA), or learned but unstructured (LoRA). ChebyMA replaces these with structured Chebyshev bases evaluated at learnable positions, introducing a polynomial inductive bias that is particularly effective when the target weight exhibits spatial smoothness.

We summarize the main contributions of this work:

\begin{itemize} \item We propose ChebyMA, a parameter-efficient weight representation grounded in Chebyshev approximation theory that models weight matrices as superpositions of polynomial hypersurfaces on learnable coordinates.

\item We establish formal approximation guarantees 
with polynomial convergence rates and prove a 
strict expressivity hierarchy across polynomial 
degrees, situating LoRA as a constrained special 
case within the framework.

\item We validate ChebyMA on four benchmarks across vision and language, demonstrating a consistently superior parameter-accuracy Pareto front, with substantial gains over LoRA, TLoRA, and 
StelLA at matched parameter budgets across all 
tested datasets. \end{itemize}

\section{Related Work}

\subsection{Parameter-Efficient Adaptation}

As pre-trained models continue to grow in scale, full fine-tuning becomes increasingly expensive in both computation and storage. Parameter-efficient adaptation methods address this by updating only a small subset of parameters while keeping the backbone frozen. Early approaches include Adapter Tuning \cite{AdapterTuning}, which inserts small bottleneck modules between transformer layers, and Prompt Tuning \cite{PromptTuning}, which prepends learnable tokens to the input sequence. While effective, these methods either modify the network architecture or alter the input space.

LoRA takes a different path by decomposing weight updates into two low-rank matrices, introducing no inference overhead since the update can be merged back into the original weights. Subsequent variants extend this idea along several axes: AdaLoRA \cite{AdaLoRA} adaptively allocates rank budgets across layers based on importance scores; QLoRA\cite{QLoRA} combines low-rank adaptation with quantization for memory efficiency; and DoRA\cite{DoRA} decouples weight updates into magnitude and direction components.

More recently, three-factor decomposition methods have emerged as a generalization of the two-factor LoRA structure. TLoRA interposes a trainable middle matrix between two fixed random projection matrices, reducing the trainable parameters to $r^2$ but at the cost of unstructured outer bases. StelLA adopts a similar three-factor form $USV^\top$ with orthogonality constraints on $U$ and $V$, ensuring the factors lie on the Stiefel manifold. LoTR \cite{LoTR} further shares the left and right factors across layers through tensor decomposition. Despite these refinements, all methods in this family operate within linear subspaces of the weight space, and their expressivity remains fundamentally bounded by the chosen rank.

\subsection{Structured Representations and Learnable Bases}
Beyond low-rank decomposition, a parallel line of research explores structured function spaces as alternatives to free-parameter representations. In weight compression, MCNC \cite{MCNC} constrains the parameter space to predefined nonlinear manifolds, while PRANC \cite{PRANC} and NOLA \cite{NOLA} project weights onto randomly selected low-dimensional subspaces. In signal representation, SIREN \cite{SIREN} parameterizes continuous fields using sinusoidal basis functions, demonstrating that structured bases can achieve high fidelity with compact parameterizations.

The idea of making basis functions themselves learnable has gained significant traction. KAN \cite{KAN} replaces fixed activation functions with learnable spline-based functions on network edges, showing that adaptive bases can outperform fixed architectures in both accuracy and interpretability. In the graph domain, ChebNet \cite{ChebNet} uses Chebyshev polynomials to approximate spectral graph filters, and Fourier Neural Operator \cite{FNO} leverages Fourier bases for learning solution operators of partial differential equations. These works share a common principle: structured, learnable function spaces offer a favorable trade-off between expressivity and parameter efficiency.
ChebyMA builds on this principle but operates at a different level. Where KAN places learnable bases on activation functions and ChebNet applies them to graph filters, ChebyMA applies Chebyshev polynomial bases directly to weight parameterization, generating weight matrices as smooth hypersurfaces evaluated on learnable coordinates.

\section{Method}
\label{sec:method}
\subsection{Chebyshev Hypersurface Parameterization}

We begin by observing that every weight matrix $W \in \mathbb{R}^{m \times n}$ can be viewed as a scalar field over its row-column index space: given a row index $i$ and a column index $j$, the entry $W(i,j)$ is a scalar value. ChebyMA approximates this scalar field through a superposition of Chebyshev polynomial hypersurfaces.

\paragraph{Single Surface.} A single hypersurface is defined by three groups of learnable parameters: a coordinate vector $u \in \mathbb{R}^m$ that maps each row index to a position in $[-1, 1]$, a coordinate vector $v \in \mathbb{R}^n$ that does the same for columns, and a coefficient matrix $A \in \mathbb{R}^{(d+1) \times (d+1)}$ that controls the shape of the surface. The Chebyshev polynomial expansion $T(\cdot)$ evaluates the first $d+1$ Chebyshev polynomials at each coordinate:

$$T(u) = \begin{bmatrix} T_0(u_1) & \cdots & T_d(u_1) \\ \vdots & \ddots & \vdots \\ T_0(u_m) & \cdots & T_d(u_m) \end{bmatrix} \in \mathbb{R}^{m \times (d+1)}$$

where $T_0(x) = 1$, $T_1(x) = x$, and $T_k(x) = 2xT_{k-1}(x) - T_{k-2}(x)$. The weight matrix generated by a single surface is then:

$$W = T(u) \; A \; T(v)^\top \in \mathbb{R}^{m \times n}$$

Each entry $W(i,j) = \sum_{p=0}^{d} \sum_{q=0}^{d} A_{pq} \, T_p(u_i) \, T_q(v_j)$ is a bivariate polynomial evaluated at the learnable coordinates $(u_i, v_j)$. The coefficient matrix $A$ encodes how different polynomial orders interact between the row and column dimensions, enabling the surface to capture non-separable row-column coupling.

\paragraph{Choice of Basis.}
The selection of Chebyshev polynomials over 
alternatives such as Legendre, Fourier, or 
monomial bases is motivated by three 
complementary properties.

First, Chebyshev polynomials are the minimax 
optimal polynomial basis on $[-1,1]$: among 
all monic polynomials of degree $d$, $T_d(x)/2^{d-1}$ 
has the smallest maximum absolute value. This 
directly translates to our setting, as the 
truncation error bound in Theorem~\ref{thm:approx} 
is an $L^\infty$ bound --- Chebyshev truncation 
minimizes the worst-case elementwise error across 
the entire weight matrix, rather than minimizing 
average error as Legendre projections would.

Second, the three-term recurrence 
$T_k(x) = 2x\,T_{k-1}(x) - T_{k-2}(x)$ 
ensures numerical stability during both forward 
evaluation and backpropagation through the 
learnable coordinates $u_k, v_k$. Monomial bases 
$\{1, x, x^2, \ldots\}$ suffer from exponential 
growth and ill-conditioning at high degree, 
leading to vanishing or exploding gradients. 
The bounded recurrence of Chebyshev polynomials 
($|T_k(x)| \leq 1$ for $x \in [-1,1]$) avoids 
this entirely.

Third, Chebyshev polynomials are orthogonal 
under the weight $1/\sqrt{1-x^2}$, which 
promotes near-independence among the columns 
of the basis matrix $T(u_k)$. This 
reduces redundancy across polynomial orders 
and improves the conditioning of the 
coefficient matrix $A_k$ during optimization. 
Fourier bases share orthogonality but require 
complex-valued computations or paired 
sine-cosine terms, doubling the parameter 
count per degree and complicating the 
real-valued einsum implementation.

\paragraph{Multi-Surface Superposition.} The full ChebyMA model superimposes $S$ such hypersurfaces, each with its own coordinate vectors and coefficient matrix:

$$W = \sum_{k=1}^{S} T(u_k) \; A_k \; T(v_k)^\top$$

Since each surface has independently learnable coordinates $u_k$ and $v_k$, their Chebyshev bases $T(u_k)$ are evaluated at different positions, making them structurally distinct. This prevents the degeneracy that occurs when multiple surfaces share fixed bases, where coefficients can be trivially merged into a single surface.

The entire forward computation is realized in a single vectorized operation:

$$W = \texttt{einsum}(\text{``smp,spq,snq} \to \text{mn"}, \; T(\mathbf{u}), \; \mathbf{A}, \; T(\mathbf{v}))$$

where $\mathbf{u} \in \mathbb{R}^{S \times m}$, $\mathbf{v} \in \mathbb{R}^{S \times n}$, and $\mathbf{A} \in \mathbb{R}^{S \times (d+1) \times (d+1)}$. No loops, no nonlinear activations between surfaces, and no change to the host network architecture. At inference time, $W$ can be materialized and cached, incurring zero additional cost.

\paragraph{Expressivity Hierarchy.} The polynomial degree $d$ controls the richness of each surface. Since the matrix $T(u_k) \, A_k \, T(v_k)^\top$ has rank at most $d+1$, a model with $S$ surfaces achieves a maximum effective rank of $S(d+1)$. This yields a strict hierarchy:

\begin{proposition}[Expressivity Hierarchy]
\label{prop:hierarchy}
For any degree $d \geq 1$, the function space of 
degree-$d$ ChebyMA strictly contains that of 
degree-$(d{-}1)$: setting all coefficients 
$A_{pq} = 0$ for $\max(p,q) = d$ recovers the 
degree-$(d{-}1)$ parameterization, while the 
converse inclusion is strict by rank argument, 
as a single degree-$d$ surface achieves rank up 
to $d+1$.

The relationship to LoRA is as follows. A single 
degree-1 ChebyMA surface generates:
$$W(i,j) = A_{00} + A_{01}\,v(j) + A_{10}\,u(i) 
+ A_{11}\,u(i)\,v(j)$$
which is a rank-2 matrix with $m + n + 4$ 
parameters. Constraining $A_{00} = A_{01} = 
A_{10} = 0$ reduces this to $W(i,j) = A_{11}
\,u(i)\,v(j)$, a rank-1 outer product identical 
to a single LoRA component. Thus LoRA is a 
constrained special case of degree-1 ChebyMA, 
and unconstrained degree-1 already provides 
twice the rank of LoRA with only 4 additional 
parameters.
\end{proposition}

\begin{algorithm}[t]
\caption{ChebyMA Forward Pass}
\label{alg:chebyma}
\textbf{Input}: Input $x \in \mathbb{R}^{b \times m}$, 
learnable parameters $\{u_k, v_k, A_k\}_{k=1}^{S}$, 
degree $d$\\
\textbf{Output}: $y \in \mathbb{R}^{b \times n}$
\begin{algorithmic}[1]
\FOR{$k = 1$ to $S$}
\STATE $T_k^u \leftarrow [T_0(u_k), \ldots, T_d(u_k)] 
\in \mathbb{R}^{m \times (d+1)}$
\STATE $T_k^v \leftarrow [T_0(v_k), \ldots, T_d(v_k)] 
\in \mathbb{R}^{n \times (d+1)}$
\ENDFOR
\STATE $W \leftarrow \texttt{einsum}(\text{smp,spq,snq} 
\to \text{mn})$
\STATE $y \leftarrow xW + b$
\STATE \textbf{return} $y$
\end{algorithmic}
\end{algorithm}

\subsection{Parameter Efficiency Analysis}

Each ChebyMA surface requires $m + n$ parameters for its coordinate vectors and $(d+1)^2$ parameters for its coefficient matrix. With $S$ surfaces, the total weight parameter count is $S\bigl(m + n + (d+1)^2\bigr)$.

To achieve the same effective rank $R$, LoRA requires $R$ rank-1 components at cost $m + n$ each, totaling $R(m+n)$ parameters. ChebyMA requires only $\lceil R/(d+1) \rceil$ surfaces, since each surface contributes rank up to $d+1$. The asymptotic parameter ratio is:

$$\frac{\text{ChebyMA}}{\text{LoRA}} = \frac{1}{d+1} + \frac{d+1}{m+n}$$

When $m + n \gg (d+1)^2$, the second term vanishes and ChebyMA uses approximately $\frac{1}{d+1}$ of LoRA's parameters for equivalent rank. At $d=2$ this yields a $3\times$ reduction. Table~\ref{tab:efficiency} confirms that the empirical ratio closely matches the theoretical prediction across a range of typical layer dimensions.

\begin{table}[t]
\centering
\begin{tabular}{lcccc}
\toprule
Layer Size & Rank & LoRA & ChebyMA ($d{=}2$) & Ratio \\
\midrule
$512 \times 128$ & 6 & 3840 & 1298 & 0.34 \\
$768 \times 256$ & 9 & 9216 & 3099 & 0.34 \\
$5000 \times 256$ & 9 & 47304 & 15795 & 0.33 \\
\bottomrule
\end{tabular}
\caption{Parameter count comparison at matched effective rank. ChebyMA consistently uses approximately one-third of LoRA's parameters, closely matching the theoretical ratio of $1/3 + (d{+}1)/(m{+}n) \approx 0.34$.}
\label{tab:efficiency}
\end{table}

\section{Theoretical Analysis}

We establish a formal approximation guarantee for ChebyMA, showing that its parameter space contains solutions capable of approximating smooth target weight matrices with controllable error.

\subsection{Approximation Expressivity Theorem}

\begin{theorem}[Approximation Expressivity Bound]
\label{thm:approx}
Let the target weight matrix $\Delta W^* \in \mathbb{R}^{m \times n}$ be generated by sampling an $r$-times continuously differentiable function $F^* \in C^r([-1,1]^2)$ at discrete coordinate points. For a single-manifold ChebyMA ($S=1$) with polynomial degree $d$, there exists a feasible solution $(\hat{A}, \hat{u}, \hat{v})$ in the parameter space such that the reconstructed weight matrix $\hat{W}$ satisfies:
$$\|\Delta W^* - \hat{W}\|_F \leq \sqrt{mn} \cdot \mathcal{O}(d^{-r})$$
\end{theorem}

The proof proceeds in four steps.

\paragraph{Step 1: Continuous Embedding.}
We assume the target matrix $\Delta W^*$ admits a continuous generating function $F^*: [-1,1]^2 \to \mathbb{R}$, such that each entry is a pointwise sample at some ideal coordinates $(x_i^*, y_j^*)$:
$$\Delta W^*_{i,j} = F^*(x_i^*, y_j^*), \quad \forall\, i \in \{1,\dots,m\},\; j \in \{1,\dots,n\}$$
Since $F^* \in C^r([-1,1]^2)$, all partial derivatives up to order $r$ exist and are bounded by a constant $M_r > 0$.

\paragraph{Step 2: Chebyshev Truncation Bound.}
By the classical theory of polynomial approximation on tensor-product domains \cite{Classic}, for any $r$-smooth function $F^*$, there exists a truncated bivariate Chebyshev polynomial of degree $d$:
$$P_d(x,y) = \sum_{p=0}^{d}\sum_{q=0}^{d} c_{pq}\, T_p(x)\, T_q(y)$$
whose uniform approximation error satisfies:
$$\|F^* - P_d\|_\infty = \max_{x,y \in [-1,1]} |F^*(x,y) - P_d(x,y)| \leq \frac{K \cdot M_r}{d^r}$$
where $K$ is a constant independent of $d$.

\paragraph{Step 3: Construction of a Feasible Solution.}
We construct a specific parameter assignment within the ChebyMA parameter space by setting the coordinate vectors to the ideal sampling points $u_i = x_i^*$, $v_j = y_j^*$, and the coefficient matrix to the Chebyshev projection coefficients $A_{pq} = c_{pq}$. Under this assignment, the ChebyMA output at each entry coincides with the truncated polynomial evaluated at the ideal coordinates:
$$\hat{W}_{i,j} = \sum_{p=0}^{d}\sum_{q=0}^{d} A_{pq}\, T_p(u_i)\, T_q(v_j) = P_d(x_i^*, y_j^*)$$
Since this is one specific feasible point in the parameter space, the globally optimal solution found by training can only achieve equal or lower error. The elementwise bound therefore holds:
$$|\Delta W^*_{i,j} - \hat{W}_{i,j}| \leq \|F^* - P_d\|_\infty \leq \frac{K \cdot M_r}{d^r}$$

\paragraph{Step 4: Frobenius Norm Aggregation.}
Summing the squared elementwise errors over all $m \times n$ entries:
\begin{align}
\|\Delta W^* - \hat{W}\|_F 
&= \sqrt{\sum_{i=1}^{m}\sum_{j=1}^{n}\left(\Delta W^*_{i,j} - \hat{W}_{i,j}\right)^2} \notag\\
&\leq \sqrt{mn} \cdot \frac{K M_r}{d^r} 
= \sqrt{mn} \cdot \mathcal{O}(d^{-r}) \notag
\end{align}

\subsection{Multi-Manifold Superposition}

The bound above establishes that a single ChebyMA surface can approximate any sufficiently smooth target weight matrix with polynomial convergence in $d$. For multi-manifold models ($S > 1$), we offer the following complementary perspective rather than imposing restrictive convergence assumptions on the residuals.

Introducing $S$ parallel manifolds expands the total parameter space dimensionality from $(d{+}1)^2$ to $S(m + n + (d{+}1)^2)$. By the intuition of Kolmogorov $n$-width theory\cite{Kolmogorov}, a higher-dimensional hypothesis space provides a tighter lower bound on the best achievable approximation error for any target in the weight space. In practice, the independently learnable coordinate systems $(u_k, v_k)$ allow different surfaces to specialize in capturing distinct structural components of the target weight matrix, such as features at different frequency bands or localized patterns in different regions of the index space. The consistent performance gains observed in our experiments as $S$ increases provide empirical support for this theoretical intuition.

\section{Experiments}

\subsection{Experimental Setup}
\paragraph{Initialization.} For ChebyMA, 
coordinate vectors $u_k$ are initialized as 
$\texttt{linspace}(-1, 1, m)$ with additive 
Gaussian noise $\mathcal{N}(0, 0.05)$, and 
$v_k$ analogously. Coefficient matrices $A_k$ 
are initialized from $\mathcal{N}(0, 0.05)$. 
This ensures the initial bases span the full 
$[-1,1]$ interval while breaking symmetry 
across surfaces. For LoRA, we follow the 
original initialization scheme. For TLoRA, 
frozen projection matrices are sampled from 
$\mathcal{N}(0, 1/\sqrt{n})$. For StelLA, 
outer matrices are initialized via QR 
decomposition of random Gaussian matrices.

\paragraph{Datasets.} We evaluate ChebyMA on four benchmark datasets across two domains. For computer vision, we use CIFAR-10 (10 classes, 50K training / 10K test images) and CIFAR-100 (100 classes, 50K / 10K) \cite{CIFAR}. For natural language processing, we use SST-2 (binary sentiment classification, 67K / 872) \cite{SST2} and AG News (4-class news topic classification, 50K / 7.6K) \cite{AGNEWS}/.

\paragraph{Architecture.} To isolate the effect of weight parameterization from other confounding factors, we adopt a consistent two-layer architecture across all experiments: a frozen pre-trained feature extractor followed by a trainable two-layer classification head. For vision tasks, we use a pre-trained ResNet-18 \cite{ResNet} as the feature extractor, producing 512-dimensional representations. For language tasks, we use a pre-trained DistilBERT \cite{DistilBERT} encoder, extracting 768-dimensional CLS token embeddings. The classification head consists of a parameterized first layer (where the five methods are compared), followed by BatchNorm\cite{BatchNorm}, ReLU\cite{ReLU}, Dropout(0.1)\cite{dropout}, and a standard linear output layer. The hidden dimensions are set to 128 for CIFAR-10 and SST-2, and 256 for CIFAR-100 and AG News.

\paragraph{Baselines.} We compare ChebyMA against four methods: (1) \textbf{Standard}: full-parameter linear layer; (2) \textbf{LoRA} \cite{LoRA}: two-factor low-rank decomposition $W = BA$; (3) \textbf{TLoRA} \cite{TLoRA}: three-factor decomposition with frozen random outer matrices; (4) \textbf{StelLA} \cite{StelLA}: three-factor decomposition with orthogonally constrained outer matrices. For each method, we sweep configurations to cover a range of parameter budgets.

\paragraph{Training Details.} All models are trained with Adam optimizer\cite{Adam}, learning rate $10^{-3}$, cosine annealing schedule, and batch size 256. Vision tasks are trained for 30 epochs; language tasks for 30 epochs. Each configuration is run 3 times with different random seeds, and we report mean accuracy and standard deviation. All experiments are conducted on a single NVIDIA T4 GPU.

\subsection{Main Results}

Table~\ref{tab:main} presents the head-to-head comparison between ChebyMA and LoRA at matched parameter budgets across all four datasets. At each operating point, we select the ChebyMA configuration and the LoRA rank whose parameter counts are closest.

\begin{table*}[t]
\centering
\begin{tabular}{llrrrr}
\toprule
Dataset & Param Range & LoRA Config & LoRA Acc (\%) & ChebyMA Config & ChebyMA Acc (\%) \\
\midrule
\multirow{4}{*}{CIFAR-10} & $\sim$650 & $r{=}1$ (640) & 40.90 $\pm$ 0.32 & $S{=}1,d{=}2$ (649) & 60.92 $\pm$ 0.20 \\
& $\sim$1300 & $r{=}2$ (1280) & 66.95 $\pm$ 0.48 & $S{=}2,d{=}3$ (1312) & 79.90 $\pm$ 0.35 \\
& $\sim$2600 & $r{=}4$ (2560) & 82.97 $\pm$ 0.17 & $S{=}3,d{=}3$ (1968) & 83.64 $\pm$ 0.12 \\
& $\sim$5200 & $r{=}8$ (5120) & 87.37 $\pm$ 0.05 & $S{=}8,d{=}3$ (5248) & 86.55 $\pm$ 0.11 \\
\midrule
\multirow{4}{*}{CIFAR-100} & $\sim$780 & $r{=}1$ (768) & 4.77 $\pm$ 0.11 & $S{=}1,d{=}3$ (784) & 17.79 $\pm$ 0.33 \\
& $\sim$1550 & $r{=}2$ (1536) & 12.14 $\pm$ 0.24 & $S{=}2,d{=}3$ (1568) & 30.18 $\pm$ 0.40 \\
& $\sim$3900 & $r{=}4$ (3072) & 28.93 $\pm$ 0.26 & $S{=}5,d{=}3$ (3920) & 44.69 $\pm$ 0.21 \\
& $\sim$6200 & $r{=}8$ (6144) & 46.96 $\pm$ 0.13 & $S{=}8,d{=}3$ (6272) & 51.01 $\pm$ 0.16 \\
\midrule
\multirow{3}{*}{SST-2} & $\sim$900 & $r{=}1$ (896) & 83.60 $\pm$ 0.25 & $S{=}1,d{=}2$ (905) & 84.29 $\pm$ 0.18 \\
& $\sim$1800 & $r{=}2$ (1792) & 83.83 $\pm$ 0.21 & $S{=}2,d{=}2$ (1810) & 84.63 $\pm$ 0.15 \\
& $\sim$2700 & $r{=}3$ (2688) & 84.17 $\pm$ 0.19 & $S{=}3,d{=}2$ (2715) & 84.94 $\pm$ 0.12 \\
\midrule
\multirow{3}{*}{AG News} & $\sim$1030 & $r{=}1$ (1024) & 84.00 $\pm$ 0.30 & $S{=}1,d{=}3$ (1040) & 89.34 $\pm$ 0.22 \\
& $\sim$2060 & $r{=}2$ (2048) & 88.18 $\pm$ 0.15 & $S{=}2,d{=}3$ (2080) & 89.47 $\pm$ 0.18 \\
& $\sim$5200 & $r{=}5$ (5120) & 89.91 $\pm$ 0.10 & $S{=}5,d{=}3$ (5200) & 89.82 $\pm$ 0.14 \\
\bottomrule
\end{tabular}
\caption{Head-to-head comparison between ChebyMA and LoRA at matched parameter budgets. Parenthesized numbers indicate exact weight parameter counts. ChebyMA outperforms LoRA at every low-to-mid parameter operating point, with the advantage most pronounced on CIFAR-100 (+18.04pp at $\sim$1550 params).}
\label{tab:main}
\end{table*}

Several patterns emerge from the results. First, ChebyMA consistently outperforms LoRA at low-to-mid parameter budgets across all four datasets. The advantage is most dramatic on CIFAR-100, where ChebyMA achieves +18.04 percentage points over LoRA at approximately 1550 parameters. Second, the gap narrows at higher parameter budgets, and LoRA occasionally matches or slightly exceeds ChebyMA when the budget approaches the full parameter count. This is expected: as the effective rank approaches full rank, the structural constraint of polynomial smoothness becomes a limitation rather than an advantage. Third, ChebyMA shows consistently smaller standard deviations, suggesting that the Chebyshev basis provides a more stable optimization landscape.

\begin{figure}[h]
\centering
\includegraphics[width=1\linewidth]{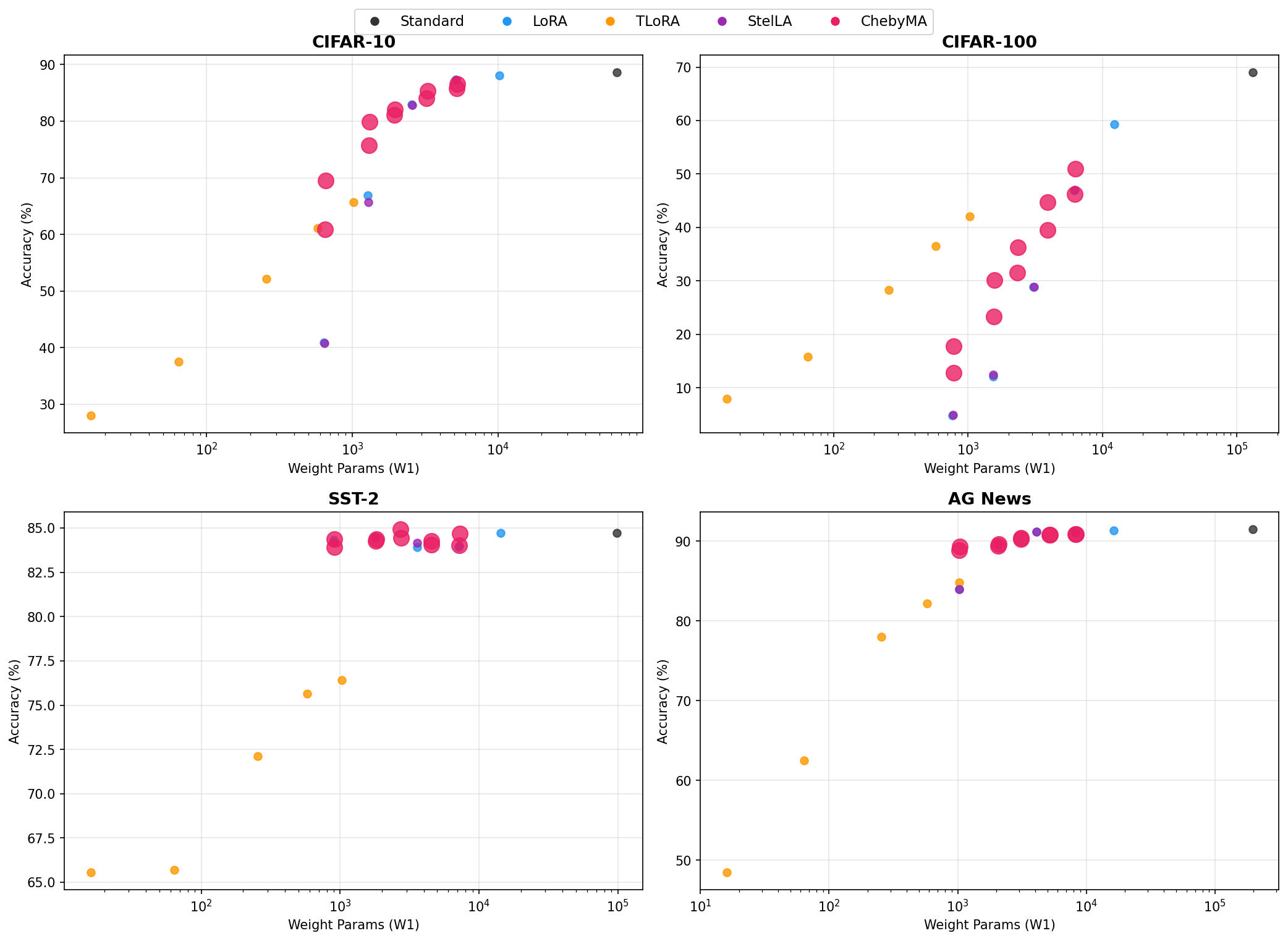}
\caption{Overview of different methods}
\label{fig:pareto}
\end{figure}

Figure~\ref{fig:pareto} visualizes the full 
parameter-accuracy Pareto front across all four datasets. 
ChebyMA consistently defines the upper-left boundary of 
the achievable trade-off at low-to-mid parameter budgets, 
while LoRA converges to comparable performance only at 
high budgets approaching the full parameter count.

\subsection{Comparison with Three-Factor Methods}

Table~\ref{tab:threefactor} compares all five methods on CIFAR-10 at representative parameter budgets. TLoRA achieves the lowest parameter counts (only $r^2$ trainable parameters) but suffers from limited expressivity due to its frozen random bases. StelLA performs comparably to LoRA, indicating that orthogonal constraints alone do not provide additional benefit. ChebyMA achieves the highest accuracy at every matched budget.

\begin{table}[t]
\centering
\begin{tabular}{lrc}
\toprule
Method & Params & Acc (\%) \\
\midrule
\multicolumn{3}{c}{\textit{$\sim$650 parameter budget}} \\
\midrule
LoRA $r{=}1$ & 640 & 40.90 \\
StelLA $r{=}1$ & 641 & 40.78 \\
TLoRA $r{=}24$ & 576 & 61.16 \\
ChebyMA $S{=}1,d{=}2$ & 649 & \textbf{60.92} \\
ChebyMA $S{=}1,d{=}3$ & 656 & \textbf{69.51} \\
\midrule
\multicolumn{3}{c}{\textit{$\sim$1300 parameter budget}} \\
\midrule
LoRA $r{=}2$ & 1280 & 66.95 \\
StelLA $r{=}2$ & 1284 & 65.75 \\
TLoRA $r{=}32$ & 1024 & 65.75 \\
ChebyMA $S{=}2,d{=}3$ & 1312 & \textbf{79.90} \\
\bottomrule
\end{tabular}
\caption{Five-way comparison on CIFAR-10 at matched parameter budgets. ChebyMA outperforms all three-factor alternatives.}
\label{tab:threefactor}
\end{table}

\subsection{Ablation Study}

We conduct ablation studies on CIFAR-10 to examine the
effect of polynomial degree $d$ and number of surfaces $S$.
Results are summarized in Table~\ref{tab:ablation}.

Increasing degree from $d{=}1$ to $d{=}3$ at fixed $S{=}1$
yields a 14.28pp improvement, consistent with the
theoretical $\mathcal{O}(d^{-r})$ convergence. Increasing
surfaces from $S{=}1$ to $S{=}8$ at fixed $d{=}3$ yields a
further 17.04pp gain with diminishing returns, as each
additional surface captures progressively finer residual
structure.

\begin{table}[t]
\centering
\begin{tabular}{ccrc}
\toprule
$d$ & $S$ & Params & Acc (\%) \\
\midrule
\multicolumn{4}{c}{\textit{Varying degree $d$ (fixed $S{=}1$)}} \\
\midrule
1 & 1 & 644 & 55.23 \\
2 & 1 & 649 & 60.92 \\
3 & 1 & 656 & 69.51 \\
\midrule
\multicolumn{4}{c}{\textit{Varying surfaces $S$ (fixed $d{=}3$)}} \\
\midrule
3 & 1 & 656 & 69.51 \\
3 & 2 & 1312 & 79.90 \\
3 & 3 & 1968 & 83.64 \\
3 & 5 & 3280 & 85.35 \\
3 & 8 & 5248 & 86.55 \\
\bottomrule
\end{tabular}
\caption{Ablation study on CIFAR-10.}
\label{tab:ablation}
\end{table}

\subsection{Discussion}

\paragraph{When does ChebyMA excel?}
ChebyMA's advantage derives from the alignment 
between its polynomial smoothness prior and the 
structure of the feature space. Pre-trained 
feature extractors such as ResNet-18 and 
DistilBERT produce embeddings where adjacent 
dimensions encode semantically related information, 
creating weight matrices with smooth spatial 
structure along the index dimensions. This 
regularity is precisely what Chebyshev bases are 
designed to exploit, explaining the large gains 
observed on CIFAR-100 and AG News. Conversely, 
on SST-2 where binary classification is nearly 
saturated by all methods, the structural prior 
offers little additional benefit. In a preliminary 
experiment where we replaced DistilBERT embeddings 
with TF-IDF bag-of-words features on AG News, 
ChebyMA's advantage disappeared entirely, 
confirming that the gain is attributable to the 
inductive bias rather than to architectural 
superiority.

\paragraph{High-parameter convergence.}
At high parameter budgets, LoRA closes the gap and
occasionally surpasses ChebyMA. On CIFAR-10, LoRA
$r{=}8$ (5120 params) reaches 87.37\% while ChebyMA
$S{=}8, d{=}3$ (5248 params) achieves 86.55\%. This is
expected: when the parameter budget is large enough to
approach full rank, the polynomial smoothness constraint
becomes restrictive rather than beneficial, as the model
needs freedom to represent non-smooth weight patterns.
This behavior is consistent with the classical bias-variance
trade-off: ChebyMA's structural prior reduces variance at
low budgets but introduces bias at high budgets.

\section{Conclusion}
\section{Conclusion}

This work demonstrates that the choice of 
parameterization for weight matrices is not merely 
an engineering decision but a question of 
approximation-theoretic optimality. By grounding 
weight representation in Chebyshev polynomial 
theory, ChebyMA reveals that the linear subspace 
assumption underlying LoRA and its variants leaves 
substantial parameter efficiency on the table. The 
consistent Pareto dominance observed across vision 
and language benchmarks suggests that structured 
function spaces offer a fundamentally better 
trade-off between expressivity and parameter count 
than unstructured low-rank decompositions, provided 
the feature space exhibits sufficient regularity.

Several limitations point to future directions. 
First, our evaluation is restricted to 
classification heads on frozen pre-trained 
features; applying ChebyMA to attention weight 
adaptation within transformer layers would test 
its effectiveness in the dominant fine-tuning 
paradigm. Second, the polynomial degree $d$ is 
currently uniform across all layers; an adaptive 
allocation mechanism that assigns higher degrees to 
layers with smoother weight structure could further 
improve efficiency. Third, combining ChebyMA with 
quantization, analogous to QLoRA, may yield 
additional memory savings. Finally, the framework 
naturally extends to higher-dimensional weight 
tensors such as convolution kernels, where 
Chebyshev hypersurfaces over four-dimensional 
index spaces may capture spatial filter structure 
more compactly than existing decomposition methods.

\bibliography{aaai2027}

\begin{thebibliography}{27}
\providecommand{\natexlab}[1]{#1}

\bibitem[{Bershatsky et~al.(2024)Bershatsky, Cherniuk, Daulbaev, Mikhalev, and Oseledets}]{LoTR}
Bershatsky, D.; Cherniuk, D.; Daulbaev, T.; Mikhalev, A.; and Oseledets, I. 2024.
\newblock LoTR: Low Tensor Rank Weight Adaptation.

\bibitem[{Dettmers et~al.(2023)Dettmers, Pagnoni, Holtzman, and Zettlemoyer}]{QLoRA}
Dettmers, T.; Pagnoni, A.; Holtzman, A.; and Zettlemoyer, L. 2023.
\newblock QLoRA: Efficient Finetuning of Quantized LLMs.

\bibitem[{He et~al.(2015)He, Zhang, Ren, and Sun}]{ResNet}
He, K.; Zhang, X.; Ren, S.; and Sun, J. 2015.
\newblock Deep Residual Learning for Image Recognition.

\bibitem[{Hinton et~al.(2012)Hinton, Srivastava, Krizhevsky, Sutskever, and Salakhutdinov}]{dropout}
Hinton, G.~E.; Srivastava, N.; Krizhevsky, A.; Sutskever, I.; and Salakhutdinov, R.~R. 2012.
\newblock Improving neural networks by preventing co-adaptation of feature detectors.

\bibitem[{Houlsby et~al.(2019)Houlsby, Giurgiu, Jastrzebski, Morrone, de~Laroussilhe, Gesmundo, Attariyan, and Gelly}]{AdapterTuning}
Houlsby, N.; Giurgiu, A.; Jastrzebski, S.; Morrone, B.; de~Laroussilhe, Q.; Gesmundo, A.; Attariyan, M.; and Gelly, S. 2019.
\newblock Parameter-Efficient Transfer Learning for NLP.

\bibitem[{Hu et~al.(2021)Hu, Shen, Wallis, Allen-Zhu, Li, Wang, Wang, and Chen}]{LoRA}
Hu, E.~J.; Shen, Y.; Wallis, P.; Allen-Zhu, Z.; Li, Y.; Wang, S.; Wang, L.; and Chen, W. 2021.
\newblock LoRA: Low-Rank Adaptation of Large Language Models.

\bibitem[{Ioffe and Szegedy(2015)}]{BatchNorm}
Ioffe, S.; and Szegedy, C. 2015.
\newblock Batch normalization: accelerating deep network training by reducing internal covariate shift.
\newblock In \emph{Proceedings of the 32nd International Conference on Machine Learning - Volume 37}, ICML'15, 448–456. JMLR.org.

\bibitem[{Islam(2026)}]{TLoRA}
Islam, T. 2026.
\newblock \emph{TLoRA: Tri-Matrix Low-Rank Adaptation of Large Language Models}, 208–223.
\newblock Springer Nature Singapore.
\newblock ISBN 9789819570782.

\bibitem[{Kingma and Ba(2014)}]{Adam}
Kingma, D.~P.; and Ba, J. 2014.
\newblock Adam: A Method for Stochastic Optimization.

\bibitem[{Kolmogorov(1936)}]{Kolmogorov}
Kolmogorov, A.~N. 1936.
\newblock {\"U}ber die beste Ann{\"a}herung von Funktionen einer gegebenen Funktionenklasse.
\newblock \emph{Annals of Mathematics}, 37(1): 107--110.

\bibitem[{Koohpayegani et~al.(2023)Koohpayegani, Navaneet, Nooralinejad, Kolouri, and Pirsiavash}]{NOLA}
Koohpayegani, S.~A.; Navaneet, K.; Nooralinejad, P.; Kolouri, S.; and Pirsiavash, H. 2023.
\newblock NOLA: Compressing LoRA using Linear Combination of Random Basis.

\bibitem[{Krizhevsky(2009)}]{CIFAR}
Krizhevsky, A. 2009.
\newblock Learning Multiple Layers of Features from Tiny Images.
\newblock 32--33.

\bibitem[{Lester, Al-Rfou, and Constant(2021)}]{PromptTuning}
Lester, B.; Al-Rfou, R.; and Constant, N. 2021.
\newblock The Power of Scale for Parameter-Efficient Prompt Tuning.

\bibitem[{Li et~al.(2020)Li, Kovachki, Azizzadenesheli, Liu, Bhattacharya, Stuart, and Anandkumar}]{FNO}
Li, Z.; Kovachki, N.; Azizzadenesheli, K.; Liu, B.; Bhattacharya, K.; Stuart, A.; and Anandkumar, A. 2020.
\newblock Fourier Neural Operator for Parametric Partial Differential Equations.

\bibitem[{Li et~al.(2025)Li, Sajadmanesh, Li, and Lyu}]{StelLA}
Li, Z.; Sajadmanesh, S.; Li, J.; and Lyu, L. 2025.
\newblock StelLA: Subspace Learning in Low-rank Adaptation using Stiefel Manifold.

\bibitem[{Liu et~al.(2024{\natexlab{a}})Liu, Wang, Yin, Molchanov, Wang, Cheng, and Chen}]{DoRA}
Liu, S.-Y.; Wang, C.-Y.; Yin, H.; Molchanov, P.; Wang, Y.-C.~F.; Cheng, K.-T.; and Chen, M.-H. 2024{\natexlab{a}}.
\newblock DoRA: Weight-Decomposed Low-Rank Adaptation.

\bibitem[{Liu et~al.(2024{\natexlab{b}})Liu, Wang, Vaidya, Ruehle, Halverson, Soljačić, Hou, and Tegmark}]{KAN}
Liu, Z.; Wang, Y.; Vaidya, S.; Ruehle, F.; Halverson, J.; Soljačić, M.; Hou, T.~Y.; and Tegmark, M. 2024{\natexlab{b}}.
\newblock KAN: Kolmogorov-Arnold Networks.

\bibitem[{Nair and Hinton(2010)}]{ReLU}
Nair, V.; and Hinton, G.~E. 2010.
\newblock Rectified linear units improve restricted boltzmann machines.
\newblock In \emph{Proceedings of the 27th International Conference on International Conference on Machine Learning}, ICML'10, 807–814. Madison, WI, USA: Omnipress.
\newblock ISBN 9781605589077.

\bibitem[{Nooralinejad et~al.(2022)Nooralinejad, Abbasi, Koohpayegani, Meibodi, Khan, Kolouri, and Pirsiavash}]{PRANC}
Nooralinejad, P.; Abbasi, A.; Koohpayegani, S.~A.; Meibodi, K.~P.; Khan, R. M.~S.; Kolouri, S.; and Pirsiavash, H. 2022.
\newblock PRANC: Pseudo RAndom Networks for Compacting deep models.

\bibitem[{Sanh et~al.(2019)Sanh, Debut, Chaumond, and Wolf}]{DistilBERT}
Sanh, V.; Debut, L.; Chaumond, J.; and Wolf, T. 2019.
\newblock DistilBERT, a distilled version of BERT: smaller, faster, cheaper and lighter.

\bibitem[{Sitzmann et~al.(2020)Sitzmann, Martel, Bergman, Lindell, and Wetzstein}]{SIREN}
Sitzmann, V.; Martel, J. N.~P.; Bergman, A.~W.; Lindell, D.~B.; and Wetzstein, G. 2020.
\newblock Implicit Neural Representations with Periodic Activation Functions.

\bibitem[{Socher et~al.(2013)Socher, Perelygin, Wu, Chuang, Manning, Ng, and Potts}]{SST2}
Socher, R.; Perelygin, A.; Wu, J.; Chuang, J.; Manning, C.~D.; Ng, A.; and Potts, C. 2013.
\newblock Recursive Deep Models for Semantic Compositionality Over a Sentiment Treebank.
\newblock In Yarowsky, D.; Baldwin, T.; Korhonen, A.; Livescu, K.; and Bethard, S., eds., \emph{Proceedings of the 2013 Conference on Empirical Methods in Natural Language Processing}, 1631--1642. Seattle, Washington, USA: Association for Computational Linguistics.

\bibitem[{Tang, Li, and Yu(2019)}]{ChebNet}
Tang, S.; Li, B.; and Yu, H. 2019.
\newblock ChebNet: Efficient and Stable Constructions of Deep Neural Networks with Rectified Power Units via Chebyshev Approximations.

\bibitem[{Thrash et~al.(2024)Thrash, Abbasi, Andreas, Nooralinejad, Koohpayegani, Pirsiavash, and Kolouri}]{MCNC}
Thrash, C.; Abbasi, A.; Andreas, R.; Nooralinejad, P.; Koohpayegani, S.~A.; Pirsiavash, H.; and Kolouri, S. 2024.
\newblock MCNC: Manifold-Constrained Reparameterization for Neural Compression.

\bibitem[{Trefethen(2019)}]{Classic}
Trefethen, L. 2019.
\newblock \emph{Approximation Theory and Approximation Practice, Extended Edition}.
\newblock Other Titles in Applied Mathematics. SIAM, Society for Industrial and Applied Mathematics.
\newblock ISBN 9781611975949.

\bibitem[{Zhang et~al.(2023)Zhang, Chen, Bukharin, Karampatziakis, He, Cheng, Chen, and Zhao}]{AdaLoRA}
Zhang, Q.; Chen, M.; Bukharin, A.; Karampatziakis, N.; He, P.; Cheng, Y.; Chen, W.; and Zhao, T. 2023.
\newblock AdaLoRA: Adaptive Budget Allocation for Parameter-Efficient Fine-Tuning.

\bibitem[{Zhang, Zhao, and LeCun(2015)}]{AGNEWS}
Zhang, X.; Zhao, J.; and LeCun, Y. 2015.
\newblock Character-level convolutional networks for text classification.
\newblock In \emph{Proceedings of the 29th International Conference on Neural Information Processing Systems - Volume 1}, NIPS'15, 649–657. Cambridge, MA, USA: MIT Press.

\end{thebibliography}


\end{document}